\documentclass{article}

\PassOptionsToPackage{numbers}{natbib}
 \usepackage[preprint]{neurips_2019}

\usepackage[utf8]{inputenc} 
\usepackage[T1]{fontenc}    
\usepackage{hyperref}       
\usepackage{url}            
\usepackage{booktabs}       
\usepackage{amsfonts}       
\usepackage{nicefrac}       
\usepackage{microtype}      
\usepackage{amsmath,amsfonts,amssymb}
\usepackage{graphicx}       
\usepackage{verbatim}

\title{Neural Assistant: Joint Action Prediction, Response Generation, and Latent Knowledge Reasoning}

\author{%
    Arvind Neelakantan\thanks{Equal contribution} \\
    Google  \\
    \texttt{aneelakantan@google.com} \\
    \And
    Semih Yavuz\footnotemark[1] \thanks{Work done when all authors were at Google}\\
    Salesforce \\
    \texttt{syavuz@salesforce.com} \\
    \And
    Sharan Narang\footnotemark[1] \\
    Google \\
    \texttt{sharannarang@google.com} \\
    \And
    Vishaal Prasad \\
    Google \\
    \And
    Ben Goodrich \footnotemark[2] \\
    Google  \\
    \And
    Daniel Duckworth  \\
    Google  \\
    \And
    Chinnadhurai Sankar \footnotemark[2] \\
    Montreal Institute for Learning Algorithms \\
    \And
    Xifeng Yan \\
    UC Santa Barbara \\
}

\begin{document}
    \maketitle
    \begin{abstract}
    Task-oriented dialog presents a difficult challenge encompassing multiple problems including multi-turn language understanding and generation, knowledge retrieval and reasoning, and action prediction. Modern dialog systems typically begin by converting conversation history to a symbolic object referred to as {\it belief state} by using supervised learning. The belief state is then used to reason on an external knowledge source whose result along with the conversation history is used in action prediction and response generation tasks independently. Such a pipeline of individually optimized components not only makes the development process cumbersome but also makes it non-trivial to leverage session-level user reinforcement signals. In this paper, we develop {\it Neural Assistant}: a single neural network model that takes conversation history and an external knowledge source as input and jointly produces both text response and action to be taken by the system as output. The model learns to reason on the provided knowledge source with weak supervision signal coming from the text generation and the action prediction tasks, hence removing the need for belief state annotations. In the MultiWOZ dataset, we study the effect of distant supervision, and the size of knowledge base on model performance. We find that the Neural Assistant without belief states is able to incorporate external knowledge information achieving higher factual accuracy scores compared to Transformer. In settings comparable to reported baseline systems, Neural Assistant when provided with oracle belief state significantly improves language generation performance.
    \end{abstract}
    
    \section{Introduction}
    \begin{figure}
        \small
        \centering
        \begin{tabular}{p{0.3\linewidth}p{0.7\linewidth}}
        	\textbf{User}  & \textbf{Neural Assistant} \\
        	\hline
        	\hline  \\ [-2ex]
        	Hello, please find me a restaurant? & \\
        	\hline  \\ [-2ex]
        	& \textbf{Response:} Sure, what type of food are you looking for? \\
        	\hline  \\ [-2ex]
        	I feel like eating mediterranean food. & \\
        	\hline  \\ [-2ex]
        	& \textbf{Response:} There are two restaurants in the moderate price range, la mimosa and shiraz restaurant. Do you have a preference? \\
        	\hline  \\ [-2ex]
        	No. & \\
            \hline  \\ [-2ex]
            & \textbf{Response:} La Mimosa is in the south part of town in the moderate price range. \\
            \hline  \\ [-2ex]
            Sounds good. Please book a table on Monday at 19:30 for 4 people. & \\
            \hline  \\ [-2ex]
            & \textbf{Response:} I have booked you a table for 4 at La Mimosa on Monday at 19:30. Your reference number is VRZRPRCM. Is there anything else I can help you with? \\
            & \textbf{Action:} \ \ restaurant-book(people=4, time=19:30, day=monday) \\
            \hline  \\ [-2ex]
            No, thanks. Have a good day. & \\
            \hline \\ [-2ex]
            & \textbf{Response:} You're welcome. Have a great day! \\
            \hline \\ [-2ex]
        \end{tabular}
        \caption{Example conversation to \textbf{book a restaurant} with the trained Neural Assistant model. The model apart from the conversation history is attending to a knowledge base containing 5K triples to incorporate external knowledge.}
        \label{fig:self-example-1}
    \end{figure}
    Task-oriented dialog systems have been a longstanding goal of artificial intelligence. A single turn of the task-oriented dialog problem can be formalized as mapping conversation history containing both user and system utterances to system action and system text response\footnote{We ignore speech-to-text and text-to-speech components in this work.}. Along with the conversation history, task-oriented systems have access to an external knowledge source (knowledge bases, documents etc.) relevant to the task that is being completed. For example, a knowledge base containing restaurants and their attributes is typically provided to a food ordering dialog system.       
    
    Dialog systems starting from the early rule-based, expert systems \cite{eliza} to the present commercially available virtual assistants like Apple Siri, Amazon Alexa, and Google Assistant rely on a pipeline containing many components. Having such a pipeline seems unavoidable given that task-oriented dialog encompasses multiple problems including multi-turn language understanding and generation, knowledge retrieval and reasoning, and action prediction. Dialog systems typically begin by converting conversation history to belief state by using supervised learning \cite{dst,dst-google,nbt,latent}. The belief state is then used to reason on an external knowledge source whose result along with the conversation history is used in action prediction and response generation tasks independently. However, relying on a pipeline of individually optimized components makes these systems hard to scale. Moreover, success of consumer facing systems rely on efficient incorporation of user reinforcement signals which is non-trivial for a pipeline system. 
    
    End-to-end learned deep learning methods have recently enjoyed much success over pipeline systems in many tasks such as image recognition, speech recognition, and machine translation \cite{deep}. Such methods have been applied to task-oriented dialog only in a limited way. For example, \citet{pipeline} use a separate deep neural network trained independently for every individual component. \citet{e2egoal} attend to a small knowledge base but do not have a generative model for text response generation. A major difficulty has been on efficiently incorporating external (structured or unstructured) knowledge to action prediction and text response generation models. In this paper, we develop {\it Neural Assistant}: a single neural network model that takes conversation history and an external knowledge source as input and jointly produces both text response and action to be taken by the system as  output. The model learns to reason on the provided knowledge source with weak supervision signal coming from the text generation and the action prediction tasks, hence removing the need for belief state annotations.  
    
    
    
    We evaluate our approach on the MultiWOZ dataset \cite{multiwoz}. The dataset contains approximately $10,000$ multi-turn dialogs between users and wizards. Along with conversations, the dataset contains both belief state and dialog act (or semantic parse) annotations. We only predict belief state annotations that correspond to action prediction and remove belief state annotations that are used for accessing the knowledge base from the dataset. We do not use the dialog act annotations in our study. Figures \ref{fig:self-example-1}, \ref{fig:self-example-2} and \ref{fig:self-example-3} are examples conversations with the Neural Assistant model. We study the effect of distant supervision, and the size of knowledge base on model performance. We find that the Neural Assistant without belief states is able to incorporate external knowledge information achieving higher factual accuracy scores compared to Transformer. In settings comparable to reported baseline systems, Neural Assistant when provided with oracle belief state significantly improves language generation performance. Even with a weakly labeled knowledge base, our system comes very close to the quality of the baseline belief state system.
    
    
    \section{Neural Assistant}
    
    \begin{figure}
        \centering
        \includegraphics[width=14cm]{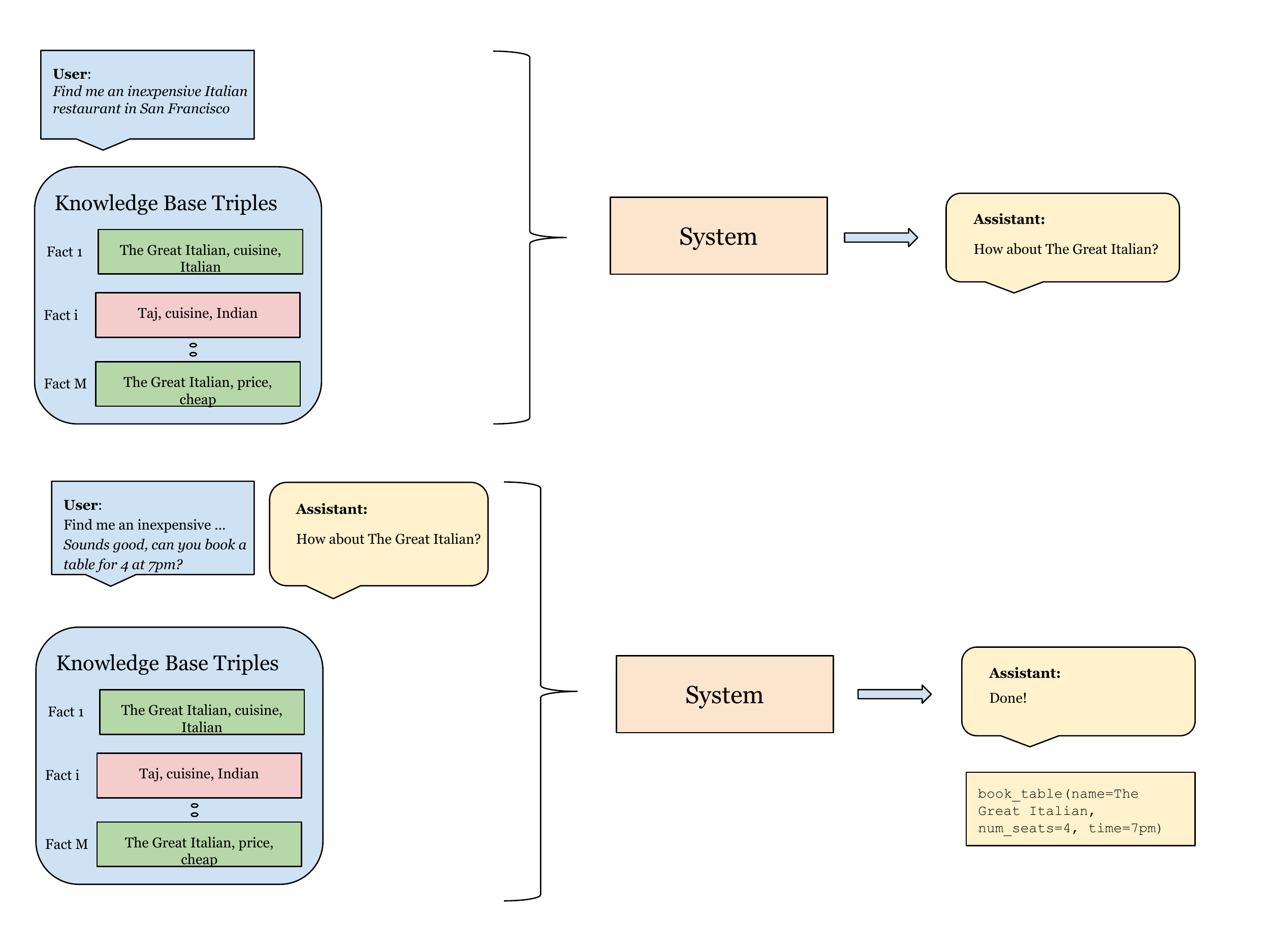}
        \caption{We formulate the task-oriented dialog problem as taking conversation history along with a relevant knowledge base (KB) as input, and generating system action and the assistant's next turn text response as output. Here we show two examples of the expected Assistant response. In the first turn no system action is taken, but in the second turn a system action is taken as all the necessary information is available. Note that only some of the triples provided in the KB are relevant to the conversation.}
        \label{fig:main}
    \end{figure}
    
    We formulate the task-oriented dialog problem as taking conversation history along with a relevant knowledge base (KB) as input, and generating system action and the assistant's next turn text response as output (Figure \ref{fig:main}). For example, the conversation history could contain a single turn of user utterance "find me an inexpensive Italian restaurant in San Francisco," and one possible next turn assistant response could be "how about The Great Italian?" Here, the external knowledge required to generate the output would be present in the provided KB. A common way to store such facts is in triple format, e.g. in this case the KB could contain (The Great Italian, type, restaurant), (The Great Italian, cuisine, Italian), (The  Great Italian, price, cheap) and so on. Given the above two utterances, the user might say "sounds good, can you book a table for 4 at 7pm?", for which the assistant performs a system action \texttt{book\_table(name=The Great Italian, num\_seats=4, time=7pm)}, and generates a text response "Done!"
    
    Neural Assistant learns to directly map the conversation history and KB to next system action and text response without any intermediate symbolic states or intermediate supervision signals. We first begin by introducing notation, then we describe the model architecture and the training objective. \\
    {\bf Conversation History} consists of alternating user and assistant turns. Let $((u^1,a^1), (u^2,a^2), \ldots, (u^U,a^U))$ denote conversation history containing $U$ turns each of user utterance ($u^i$)  and assistant utterance ($a^i$). The user and assistant turns each contain variable number of word tokens. \\ 
    {\bf Knowledge Base:} We assume the external knowledge required to solve the task is provided. While it is possible to leverage both structured and unstructured knowledge in our framework, in this work, we consider external knowledge in the form of structured KB containing a list of triples. Let $ K=(e^1_1, r_1, e^2_1), (e^1_2, r_2, e^2_2), \ldots, (e^1_M, r_K, e^2_M)$ be the list of triples in the provided KB. \\
    {\bf Output} consists of both the system action and text response.

    \subsection{Model and Training Objective}
    \label{section:attention}
    \begin{figure}
        \centering
        \includegraphics[width=14cm]{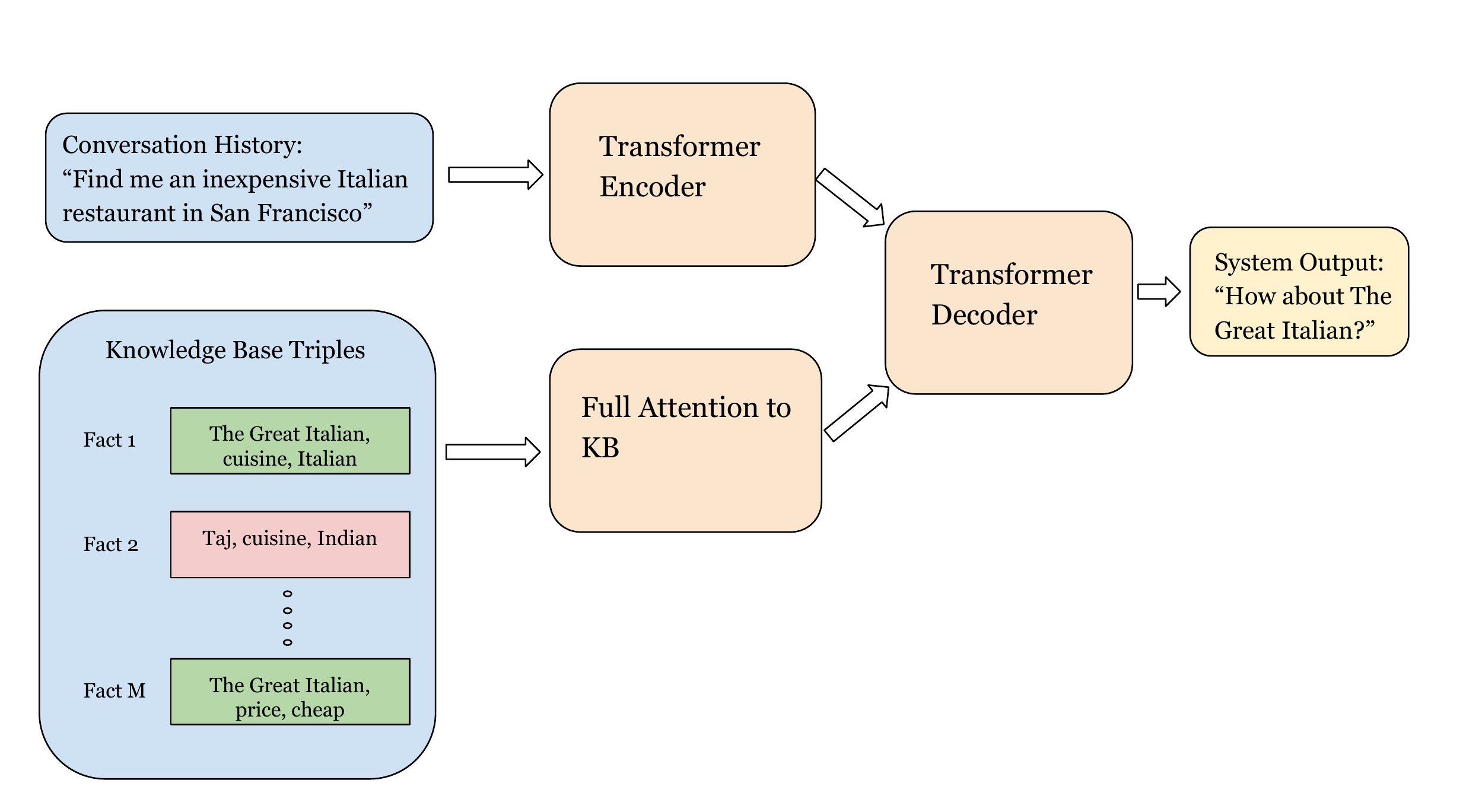}
        \caption{Neural Assistant model with attention to provided knowledge base. Transformer encoder consumes the conversation history (containing alternating user and assistant turns). The Transformer decoder generates the output sequence after performing decoder attention on the encoded conversation history and the knowledge base. Note that only some of the triples provided in the KB are relevant to the conversation and the model has to learn to pick them from weak supervision signal.}
        \label{fig:sparse}
    \end{figure}

    Neural Assistant is an extension of the Transformer \cite{transformer} encoder-decoder model. Our model additionally attends to the provided KB to incorporate external knowledge. We encode the knowledge triples separately (in parallel) and the decoder attends to the triples in addition to the input conversation history.  
    
    Transformer encoder is used to consume the input conversation history. Let $x=(x_1,x_2,\ldots, x_P)$ be the concatenated conversation history (both assistant and user turns separated by delimiters) containing $P$ tokens. Then the encoder produces $P$ hidden states $h_1, h_2, \ldots, h_P$ after word embedding lookup and multiple self attention layers. We represent each KB fact as an average of the word embeddings of the tokenized triple. We denote the representations of the triples  $ K=(e^1_1, r_1, e^2_1), (e^1_2, r_2, e^2_2), \ldots, (e^1_M, r_K, e^2_M)$ by $v_1, v_2, \ldots, v_M$. 
    
    The transformer decoder which contains both self-attention and encoder-decoder attention layers generates the output sequence consisting of both the system action and text response one token at a time, left-to-right. We  tokenize the system action with the text tokenizer and generate a concatenated version of system action and text response as one long sequence. While the encoder-decoder attention layers in Transformer \cite{transformer} only attend to input (conversation history), we make a modification to the Transformer decoder where it attends to both the encoder hidden states of the conversation history, and to the representation of the fact triples (Figure \ref{fig:sparse}). So, the decoder attention heads attend to the set $[h_1, \ldots, h_P, v_1, \ldots, v_M]$. In previous work with Transformer, the decoder attends only to $[h_1, \ldots, h_P]$. 
    
    Let $y=(y_1,y_2, \ldots, y_T)$ denote the target sequence, we model the target sequence distribution as
    \begin{equation}
    \label{eq:gen}
        P_{gen}(y | x,K) = \prod_{t=1}^{T} P_\theta(y_t | y_{1:t-1}, x,K).
    \end{equation}

    Given a training set of $N$ examples $((x^1, K^1, y^1), (x^2, K^2, y^2), \ldots, (x^N, K^N, y^N))$, the objective function to be maximized is given by
    \begin{equation}
    \label{eq:gen-loss}
        \mathcal{L}_{\text{gen}}(\theta)
        = \sum_{i=1}^N \sum_{t=1}^{T_i} \log p_{\theta}(y^i_t | y_{1:t-1}^i, x^i, K^i).
    \end{equation}
    We use teacher-forcing \citep{williams1989learning} where the model conditions on ground-truth previous tokens in the output and ground-truth previous assistant turns in the conversation history.
    
    \subsection{Distant Supervision}
    \label{section:distant}
    We adopt a technique  called distant supervision \cite{distant} widely used in knowledge base construction research. At train time, we (weakly) label facts in the KB positive if some word in the entities of the triple $(e^1, e^2)$ in $(e^1, r, e^2)$ are present in the ground-truth target sequence. This weak supervision signal could potentially guide the decoder attention to KB described above. 
    
    The distant supervision objective to be maximized is given by
    \begin{equation}
    \label{eq:d-loss}
        \mathcal{L}_{\text{d}}(\theta)
        =  \sum_{i=1}^N \sum_{m=1}^{M_i} ( \log q_m [[y_m == 1]] +  \log (1 - q_m) [[y_m == 0]])
    \end{equation}
    where $q_m$ is the attention probability, and $y_m$ is an indicator variable that is set to 1 if some word in the entities of the triple are present in the ground-truth target sequence and 0 otherwise. The model now maximizes an interpolation of the two objective functions in Equation \ref{eq:gen-loss} and Equation \ref{eq:d-loss}, given by
    \begin{equation}
    \label{eq:final}
        \mathcal{L}_{\text{final}}(\theta)
        =  \alpha  \mathcal{L}_{\text{gen}}(\theta) + (1-\alpha) \mathcal{L}_{\text{d}}(\theta).
    \end{equation}
    where $\alpha \in [0, 1]$ is a weighting term tuned on the development set.
    
    \section{Related Work}
    In past work, dialog systems have generally relied on pipeline systems \cite{singh,full}. Deep learning has been applied to task-oriented dialog in many recent studies \cite{dst,sclstm,hcn,dst-google,nbt,latent,e2egoal}. One line of work has been on using deep learning to predict belief states using supervised learning \cite{dst,dst-google}. The other line of work makes use of pipelines consisting of many components each represented as a neural network trained independently \cite{sclstm,nbt,latent,copy-dialog}. 
    
     The line of work closest to our is in the use of memory networks \cite{memory-net,e2e-memory-net} for task-oriented dialog \cite{e2egoal,gated,qrn}. While all these works incorporate an external knowledge source directly to text response generation, they do not employ a generative model for response generation, and instead rely on selecting a response from a list of candidate responses which is impractical in real-word settings. More recently, \citet{global} use a generative model instead of a text classification model but they along with previous work \cite{e2egoal,gated,qrn} work with much smaller knowledge bases where unlike in our case, full softmax attention over the knowledge base is computationally feasible. Also, they do not  generate both the text response and system action jointly together in a single model.

    Other kinds of dialog tasks have also been tackled by deep learning. This line of work has predominantly been in the chit-chat setting where generative deep learning models are used to generate text responses \cite{conv-seq2seq,hierarchical,diverse}. More recent work has extended this line of work to language based negotiation games \cite{deal} and dialog systems with persona \cite{persona}.  
    
    \section{Experiments}
    \label{sec:experiments}

    We evaluate our method on the MultiWOZ \cite{multiwoz} dataset. The dataset contains close to $8,000$ training examples and $1,000$ examples in both the validation and test sets. We report results on test set in the tables below. The dataset includes an associated knowledge base containing $28,483$ triples. To evaluate the performance of different methods, we use F-1 score for action prediction (Action F-1) and BLEU score for text response generation. Apart from  BLEU score which primarily measures fluency, we also report Entity-F1 score which is an approximate metric to measure the factualness of the text response. We get the list of entities mentioned in the ground truth response and compare it to the list of entities in the model prediction. We use exact string match to get the list of entities.  Our models are implemented in the Tensor2Tensor \cite{tensor2tensor} framework. All models are trained for 50k steps. Due to the small size of the dataset, we use the \textit{tiny} Transformer hyper-parameter setting in Tensor2Tensor. Unless otherwise stated the Neural Assistant is trained without the distant supervision objective. 
     
    Figures \ref{fig:self-example-1}, \ref{fig:self-example-2} and \ref{fig:self-example-3} are examples conversations with the Neural Assistant model in real-time to complete a task. Note that the model is trained at turn-level, where the dialog history fed into model as input consists of the previous ground-truth turns of the dialog example. The model is not exposed to text responses it generated in the previous turns as a part of input dialog history in training time. However, in the conversations in the figures, the actual text responses generated by model itself are used as the assistant's side of dialog history to be fed as input to model for generating text responses and actions in the following turns of the dialog.
     
    \subsection{Results}
    
    First, we benchmark the Transformer model on belief state prediction and text generation problems to compare with the results reported in \citet{multiwoz}. The Transformer baseline models only take the conversation history as input. They skip the KB and do not use oracle belief state annotations. The text generation results are in Table \ref{tab:baseline-results}. We treat belief state prediction also as a sequence-to-sequence problem and achieve $72.9$ F-1 score on belief state prediction, which is once again significantly higher than $63.8$ F-1 score from the baseline system.      
    
    Next, we start reporting results on the Neural Assistant model. We evaluate our framework in increasingly harder settings by gradually increasing the size of the external knowledge source to be incorporated by the model.  To begin with as done in \citet{multiwoz}, we include oracle belief state annotations which reduces the size of the KB to be considered for a given input to be less than 10 triples. As shown in Table \ref{tab:baseline-results}, the Neural Assistant model achieves a BLEU score of $25.71$, significantly higher than the baseline system \cite{multiwoz} that gets $18.9$ BLEU score. Since the oracle belief states are provided to the model, we do not evaluate the Entity F-1 and Action F-1 score for this setting. Then we make the setting slightly harder where the model consumes only weakly labeled positive triples from distant supervision (Section \ref{section:distant}). Here, the size of the KB to be considered is around 50 triples per example. Even with a weakly labeled knowledge base, our system comes very close to the quality of the baseline belief state system.

    
    %

     \begin{table}
	    \centering
	    \begin{tabular}{l| c c c}
	    \toprule
	    Model & BLEU & Action F-1 & Entity F-1 \\ [1.2mm]
	    \midrule
	    System with Oracle Belief State \cite{multiwoz}  & $18.9$ & N/A & N/A  \\  [1.2mm]
	     Transformer & $14.1$ & $90.0$ &  $40.0$\\ [1.2mm] \hline
	     & & & \\
	    Neural Assistant (oracle triples) & $\mathbf{25.71}$ & N/A & N/A \\  [1.2mm]
	    Neural Assistant (weakly labeled positive triples) & $17.9$ & $\mathbf{90.8}$ & $\mathbf{90.9}$ \\  [1.2mm]
	    \bottomrule
	    \end{tabular}
	    \vspace{2mm}
	    \caption{Comparison of Neural Assistant with other baselines.}
	    \label{tab:baseline-results}
	\end{table}
     
     \subsection{Neural Assistant with Large Knowledge Base}
      Now, we carefully study the extent to which Neural Assistant models can handle large KBs. We get the set of weakly labeled positive triples for every example and fill up the rest of KB with randomly sampled negative examples both at train and test time. The goal of this experiment is to study the effect of KB size on Neural Assistant performance. Another way to look at this experiment is to study the extent to which our model can tolerate the errors of a retrieval system. The performance of Neural Assistant on different KB sizes are in Table \ref{table:kb_and_ds_results}. The BLEU score and Entity F-1 scores for Neural Assistant reduce as the KB size increases. The model is able to incorporate external knowledge effectively as long as the KB size is $2000$ triples or smaller. Beyond that, the Entity F-1 score degrades quite rapidly. We also study the effect of distant supervision discussed in Section \ref{section:distant} as an additional training objective on Neural Assistant performance. Our experiments show that in some cases distant supervision helps the model to get better performance particularly higher entity F-1 score but not in all cases. Finally, we report results from using the entire KB at test time using a model that is trained with $5,000$ triples at train time without distant supervision. In this setting, the entity F-1 score is quite low indicating since the model is not able to select the relevant entities from the knowledge base at test time. The model cannot consume the entire KB at train time as it runs out of memory on ML accelerators.

    \begin{table}
    	\centering
    		\begin{tabular}{l | c c c | c c c}
    		\hline
    		\toprule
    		 & \multicolumn{3}{c|}{ Neural Assistant}   & \multicolumn{3}{c}{Neural Assistant with DS}  \\
    		\midrule
            Size of KB & BLEU & Action F-1 & Entity F-1  & BLEU & Action F-1 &  Entity F-1 \\[1.2mm]   
            \midrule
			100 & $16.7$ & $87.6$ & $74.0$ & $17.0$ & $87.2$ & $74.1$  \\ [1.2mm]
            2000 & $15.5$ & $87.9$ & $65.0$ & $15.4$ & $86.3$ & $65.5$  \\[1.2mm]
			5000 & $14.3$ & $86.2$ & $59.2$ & $14.4$ & $85.7$ & $59.0$  \\[1.2mm]
			8000 & $13.9$ & $85.7$ & $55.5$ & $14.0$ & $88.6$ & $54.9$   \\[1.2mm]
			12000 & $14.3$ & $85.5$ & $50.3$ & $13.7$ & $87.7$ & $50.2$  \\[1.2mm]
			13000 & $13.7$ & $88.1$ & $48.0$ & $13.8$ & $86.7$ & $49.2$   \\ [1.2mm]  \hline
			& &  & & & & \\
			28483 (at test)  & $13.5$ & $90.2$ & $42.9$ & - & - & - \\[1.2mm]
			\bottomrule
			
    	    \end{tabular}
    	    \vspace{2mm}
    	\caption{We compare the performance of Neural Assistant with and without distant supervision (Section \ref{section:distant}). The performance of the Neural Assistant model drops as the number of negative examples (Size of KB) increases. Neural Assistant is able to incorporate external knowledge information especially when the size of KB is not too large.}
	\label{table:kb_and_ds_results}
	\end{table}

    \section{Conclusion}
     We introduced {\it Neural Assistant}, a single neural network model that takes conversation history and an external knowledge source as input and jointly produces both text response and action to be taken by the system as output. The model learns to reason on the provided knowledge source with weak supervision signal coming from the text generation and the action prediction tasks, thereby removing the need for belief state annotations. Future work involves developing techniques that can effectively handle much larger KBs.
    
    \begin{figure}
        \centering
        \small
        \begin{tabular}{p{0.3\linewidth}p{0.7\linewidth}}
        	\textbf{User}  & \textbf{Neural Assistant} \\
        	\hline
        	\hline  \\ [-2ex]
        	Hi there, can you find me a place to stay? & \\
        	\hline  \\ [-2ex]
        	& \textbf{Response:} Sure, do you have a price range in mind?  \\
        	\hline  \\ [-2ex]
        	I prefer a place with moderate price. & \\
        	\hline  \\ [-2ex]
        	& \textbf{Response:} There are several in the moderate price range. Do you have a preference? \\
        	\hline  \\ [-2ex]
        	I would prefer it to be in the south. & \\
            \hline  \\ [-2ex]
            & \textbf{Response:} I have two guesthouses in the south in the moderate price range. Would you like to book one of them? \\
            \hline  \\ [-2ex]
            Yes, please. & \\
            \hline  \\ [-2ex]
            & \textbf{Response:} I'd be happy to book you a room at aylesbray lodge guest house, and bridge guest house. Which one do you prefer? \\
            \hline  \\ [-2ex]
            bridge guest house sounds good. & \\
            \hline \\ [-2ex]
            & \textbf{Response:} I can book that for you. How many people will be staying and how many nights? \\
            \hline \\ [-2ex]
            4 people will be staying for 2 nights starting on this friday. & \\
            \hline  \\ [-2ex]
            & \textbf{Response:} Booking was successful. Reference number is : GBHPGZH. \\
            & \textbf{Action:} \ \ hotel-book(stay=2,people=4,day=friday) \\
            \hline  \\ [-2ex]
            That was quick, thank you Neural Assistant! & \\
            \hline  \\ [-2ex]
            & \textbf{Response:} You're welcome. Is there anything else I can help you with? \\
            \hline  \\ [-2ex]
            No, thanks. Have good week! & \\
            \hline  \\ [-2ex]
            & \textbf{Response:} You're welcome. Have a great day. \\
            \hline  \\ [-2ex]
        \end{tabular}
        \caption{Example conversation to \textbf{book a hotel}  with the trained Neural Assistant model. The model apart from the conversation history is attending to a knowledge base containing 5K triples to incorporate external knowledge.}
        \label{fig:self-example-2}
    \end{figure}

    \begin{figure}\
        \centering
        \small
        \begin{tabular}{p{0.3\linewidth}p{0.7\linewidth}}
        	\textbf{User}  & \textbf{Neural Assistant} \\
        	\hline
        	\hline  \\ [-2ex]
        	Hello, can you find me a hotel to stay? & \\
        	\hline  \\ [-2ex]
        	& \textbf{Response:} Sure, do you have a price range in mind?  \\
        	\hline  \\ [-2ex]
        	Yes, somewhere in the moderate price range. & \\
        	\hline  \\ [-2ex]
        	& \textbf{Response:} There are several in the moderate price range. Do you have a preference on area? \\
        	\hline  \\ [-2ex]
        	Yes, how about somewhere in the north? & \\
            \hline  \\ [-2ex]
            & \textbf{Response:} I have two options for you. The Ashley Hotel and Lovell Lodge. Would you like to book one of these? \\
            \hline  \\ [-2ex]
            Yes. & \\
            \hline  \\ [-2ex]
            & \textbf{Response:} What day would you like to start your stay? \\
            \hline  \\ [-2ex]
            Can you book it for 1 people starting on Thursday for 2 night? & \\
            \hline \\ [-2ex]
            & \textbf{Response:} Booking was successful. Reference number is : FKQVSY9Y. \\
            & \textbf{Action:} \ \ hotel-book(stay=2,people=1,day=thursday) \\
            \hline \\ [-2ex]
            Thank you! & \\
            \hline  \\ [-2ex]
            & \textbf{Response:} You're welcome. Is there anything else I can help you with? \\
            \hline  \\ [-2ex]
            Can you also find me a nice restaurant? & \\
            \hline  \\ [-2ex]
            & \textbf{Response:} Sure, do you have a food preference? \\
            \hline  \\ [-2ex]
            Yes, I prefer Italian food. & \\
            \hline  \\ [-2ex]
            & \textbf{Response:} There are two Italian restaurants in the south area. One is in the moderate price range and one is in the centre. Do you have a preference? \\
            \hline  \\ [-2ex]
            I prefer the one in moderate price range. & \\
            \hline  \\ [-2ex]
            & \textbf{Response:} Pizza Hut Cherry Hinton is in the south part of town and is in the moderate price range. Would you like to book a table? \\
            \hline  \\ [-2ex]
            Yes, please book a table for 1 person at 20:30. & \\
            \hline  \\ [-2ex]
            & \textbf{Response:} I'm sorry, that time is not available. Would you like to try another time or day? \\
            & \textbf{Action:} \ \ restaurant-book(people=1,time=20:30,day=thursday) \\
            \hline  \\ [-2ex]
            Try 19:30. & \\
            \hline  \\ [-2ex]
            & \textbf{Response:} I was able to book you at Pizza Hut Cherry Hinton. Your reference number is FG1KWP5. Is there anything else I can help you with? \\
            \hline  \\ [-2ex]
            No, thanks. Have a good night. & \\
            \hline  \\ [-2ex]
            & \textbf{Response:} Thank you for using our service. Have a great day. \\
            \hline  \\ [-2ex]
        \end{tabular}
    \caption{Example conversation to \textbf{book a hotel and a restaurant} in a single conversation with the trained Neural Assistant model. The model apart from the conversation history is attending to a knowledge base containing 5K triples to incorporate external knowledge.}
        \label{fig:self-example-3}
    \end{figure}

    \bibliographystyle{plainnat}
    \bibliography{references}
\end{document}